\documentclass[10pt,twocolumn,letterpaper]{article}

\usepackage{cvpr}
\usepackage{times}
\usepackage{epsfig}
\usepackage{graphicx}
\usepackage{amsmath}
\usepackage{amssymb}
\usepackage{booktabs}
\usepackage[normalem]{ulem}
\usepackage{xcolor}
\usepackage{tikz}
\usepackage{textcomp}
\usepackage{multirow}

\usepackage[pagebackref=true,breaklinks=true,letterpaper=true,colorlinks,bookmarks=false]{hyperref}

\newcommand{\new}[1]{#1}

\cvprfinalcopy 

\ifcvprfinal\fi
\begin{document}

\title{Viraliency: Pooling Local Virality\vspace{-6mm}}

\author{Xavier Alameda-Pineda$ ^{1,2}$, Andrea Pilzer$^1$, Dan Xu$^1$, Nicu Sebe$^1$, Elisa Ricci$^{3,4}$\\
$^1$ University of Trento, $^2$ Perception Team, INRIA Grenoble, $^3$  University of Perugia, $^4$ Fondazione Bruno Kessler\\
{\tt\small xavier.alameda-pineda@inria.fr, \{andrea.pilzer,dan.xu,nicu.sebe\}@unitn.it, elisa.ricci@diei.unipg.it}
}

\newcommand{\todo}[1]{\textcolor{red}{#1}}

\maketitle

\begin{abstract}
In our overly-connected world, the automatic recognition of virality -- the quality of an image or video to be rapidly and widely spread in 
social networks -- is of crucial importance, and has recently awaken the interest of the computer vision community. Concurrently, recent progress in 
deep learning architectures showed that global pooling strategies allow the extraction of activation maps, which highlight the parts of the image 
most likely to contain instances of a certain class. We extend this concept by introducing a pooling layer that learns the size of the support 
area to be averaged: the learned top-$N$ average (LENA) pooling. We hypothesize that the latent concepts (feature maps) describing 
virality may require such a rich pooling strategy. We assess the effectiveness of the LENA layer by appending it on top of a convolutional 
siamese architecture and evaluate its performance on the task of predicting and localizing virality. We report experiments on 
two publicly available datasets annotated for virality and show that our method outperforms state-of-the-art approaches.\vspace{-5mm}
%
%
%
\end{abstract}

\section{Introduction}
Beyond the automatic understanding of objective properties of images, such as the presence of an object and its position 
in the scene, the computer vision community also invested efforts in analyzing subjective attributes of visual data. 
Memorability~\cite{celikkale2013visual,isola2014makes}, popularity~\cite{khosla2014makes}, 
virality~\cite{deza2015understanding} and emotional content~\cite{alameda2016recognizing,peng2015mixed} are few examples 
of such attributes. Further analysis was conducted to understand which parts of the image were responsible for the 
recognition of such properties. For instance, Doersch~\etal identified specific mid-level visual patterns when 
recognizing city-based architectural styles~\cite{doersch2012makes}. De Nadai~\etal studied the perception of safety in 
urban scenes~\cite{de2016safer}, detecting which areas in an image are responsible for this perception. Naturally, many 
researchers also wondered how to transform an image so as to enhance or diminish its subjective properties, or even how 
to generate images with such properties. In this regard, Koshla~\etal~\cite{khosla2013modifying} investigated how to 
transform a face image so as to make it more memorable. Given a natural image, Gatys~\etal~\cite{gatys2015neural} showed 
how to generate a stylized image from a natural image and an artistic painting using deep neural architectures.

\begin{figure}
 \centering
 \includegraphics[width=\linewidth]{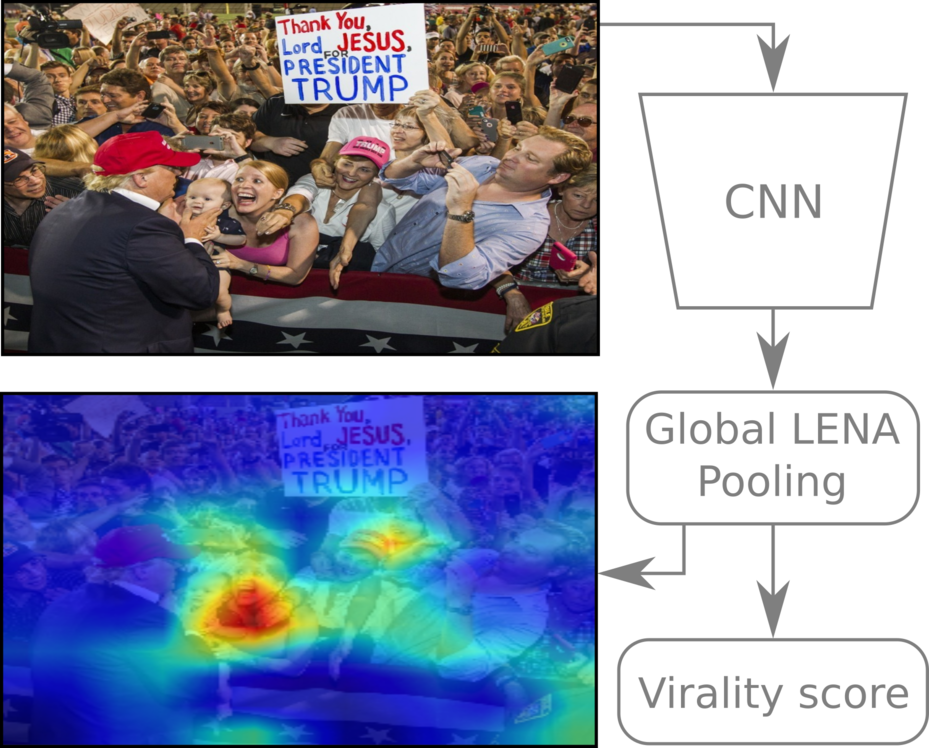}
\caption{An image related to the U.S. presidential elections that went viral in different social networks, and its 
estimated viraliency map. The proposed pipeline: (i) A convolutional deep architecture is trained to generate 
virality-sensitive feature maps of the image; (ii) These features are passed through a LENA global pooling layer; (iii) 
The global pooling provides activations to estimate the virality score as well as a rough localization of the virally 
salient parts of the image, hence the title \textit{viraliency}.}\vspace{-5mm}
\label{approach}
\end{figure}

The particular case of virality -- the quality of an image or video to be rapidly and widely spread on social networks-- 
is of crucial importance in our overly-connected world, and it is the focus of this study.
%
%
%
%
%
We hypothesize that, within an image, there are few different virally salient regions, \ie areas responsible for making 
the image viral. Inspired from previous research studies~\cite{zhou2015learning,porzi2015predicting}, 
we introduce a novel global pooling layer, the learned top-$N$ average (LENA) pooling layer, specifically designed to 
automatically detect the visual patterns correlated with virality, \ie the \textit{viraliency map} 
(Figure~\ref{approach}). We further show that, by embedding our LENA pooling into a convolutional deep architecture, we 
can successfully predict the virality score of an image and simultaneously uncover its viraliency map. We test the LENA 
layer within different network architectures and perform an extensive experimental evaluation on two recent and 
publicly available datasets used for visual virality analysis~\cite{guerini2013exploring,deza2015understanding}, 
demonstrating that our method outperforms state-of-the-art approaches on virality prediction and localization.

Up to the authors' knowledge, this is the first study addressing the complex task of recognizing image virality with an 
end-to-end trainable deep network. The proposed architecture is specifically designed to simultaneously predict image 
virality and to automatically identify the parts of the image responsible for it \new{(without using any information on 
virality localization)}. Secondly, we introduce the LENA pooling layer and demonstrate its effectiveness in virality 
prediction and in enhancing the identification of virally salient zones (viraliency maps) in two publicly available 
datasets. Interestingly, we also show that including objectness maps derived from pretrained deep models is advantageous 
for the task of interest. Finally, we complement the existing datasets with virality localization annotations and 
provide visualization results for the intuitive understanding of the advantage of the LENA pooling layer.


\section{Related work}
Our work is closely related to two recent trends in the computer vision community: (i) the understanding and recognition of subjective properties of 
visual data 
and (ii)~~the use of global pooling layers in deep neural networks for weakly-supervised localization.

Understanding and recognizing subjective properties of images is challenging because, unless 
some related information can be
automatically extracted from auxiliary data sources (\ie metadata), 
collecting and annotating datasets is a tremendous effort. Indeed, given that the perception of 
subjective properties inherently depends on the perceiver in a strong manner, each image requires to be annotated by a relatively large 
number of people. Such strategy could be an option if a web-based platform already exists and it provides annotations, as for instance for 
aesthetics~\cite{dhar2011high}. Usually, it is easier to give relative annotations between a pair of images than absolute scores. 
This scheme has been successfully employed in the past for urban perception~\cite{dubey2016deep} and emotion recognition from abstract 
paintings~\cite{sartori2015who}, but typically requires some post-processing to handle noisy annotations.

This problem is aggravated by the data-hungry deep neural architectures. It is therefore unsurprising 
that the computer vision community payed special attention to those subjective properties for which semi-automatic annotation schemes can be
devised. Memorability~\cite{isola2011makes,khosla2012memorability,celikkale2013visual,khosla2013modifying,isola2014makes} is the example \textit{par 
excellence}, since the memory game sets a very user-friendly and enjoyable environment for memorability annotation. Popularity and virality fall also 
into this category, thanks to the computational proxies provided by social networks. In particular, statistics of \textit{upvotes}, \textit{likes}, 
\textit{shares} and \textit{resubmissions} can provide almost-clean labeled datasets. The difference between virality and popularity is that 
viral images have been upvoted/liked and have been 
shared/resubmitted several times, while popular images do not satisfy the latter, as reported in~\cite{deza2015understanding}. 
Khosla~\etal~\cite{khosla2014makes} analyzed the 
popularity of images in Flickr. The study from 
Deza and Parikh~\cite{deza2015understanding} was the first attempt to understand virality from visual 
content by focusing on the mid-level attributes of images. In this paper we explore 
an orthogonal research direction to~\cite{deza2015understanding} and propose a deep architecture including a novel pooling layer specifically designed to
understand which parts of an image contribute to virality. To our knowledge, this is the first work focusing on this
aspect.

Our proposal is inspired from recent research on deep networks analyzing the role of global pooling layers
for weakly-supervised object localization \cite{oquab2015object,zhou2015learning}. As for the case of subjective 
properties, collecting datasets with annotations of the objects' bounding box is very tedious. Therefore, 
researchers in computer vision found alternative ways to tackle the problem of detection using only image-level annotations, 
\ie simply indicating the object presence/absence in the image~\cite{cinbis2014multi,bilen2015weakly}. A recent line of 
research explored weakly-supervised object localization through the use of global pooling layers on convolutional neural networks. 
For instance, Oquab \etal~\cite{oquab2015object} analyzed the ability of global max pooling to predict locations of objects
inside a deep network trained for object classification. Similarly, Zhou \etal~\cite{zhou2015learning} 
addressed weakly-supervised object localization using 
global average pooling and extended their analysis to abstract concepts. Porzi \etal\cite{porzi2015predicting}
introduced the top-$N$ average pooling to study subjective judgments from urban scenes and automatically extract image regions
responsible for these judgments. In this paper we follow this research direction and analyze if global pooling layers
are also effective when used for classifying and localizing patterns associated to virality. In addition, we step beyond previous research 
studies by introducing the \textit{learned} top-$N$ average pooling, able to learn the size of the support area to be averaged. LENA is designed to 
find 
the best compromise between average pooling and max pooling, and it is described in the next section\new{, together with the overall proposed 
architecture}.

\vspace{-1mm}
\section{Viraliency through global LENA pooling}
\vspace{-1mm}
\subsection{Predicting and localizing virality}

\begin{figure}
 \centering
 \includegraphics[width=\columnwidth]{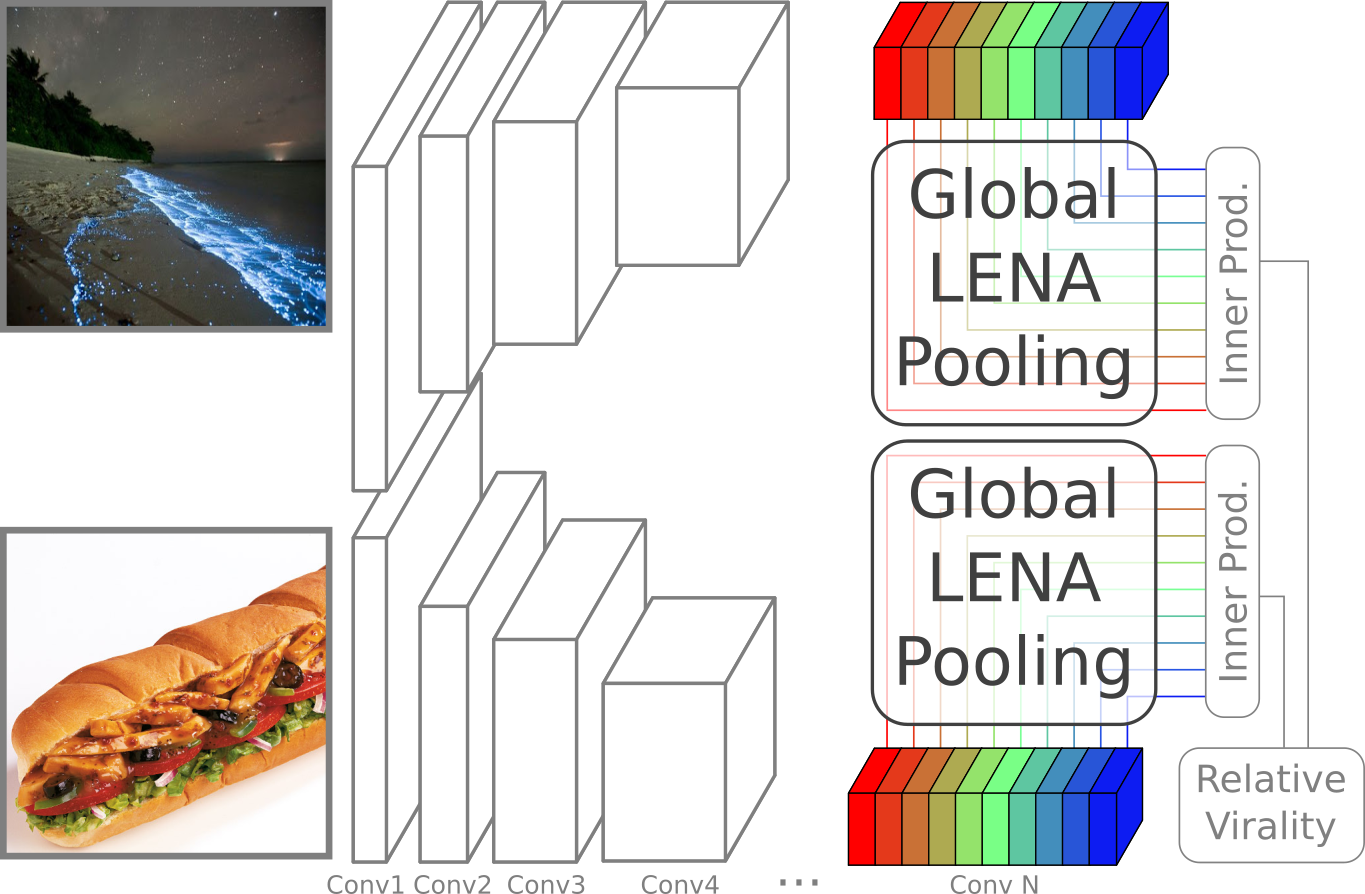}
\caption{The proposed end-to-end trainable siamese architecture consisting on: (i) a fully convolutional front-end, 
(ii) a global learned top-$N$ average (LENA) pooling (used on top of the convolutional structure to extract the 
activation from each feature map) and (iii) an inner product layer to estimate the relative virality between two images 
from the extracted activations.\label{fig:architecture}\vspace{-3mm}}
\end{figure}

We use an end-to-end trainable siamese deep neural network consisting on three main blocks: a fully 
convolutional front-end, a global LENA pooling layer and a final inner-product layer used to predict the virality score. These three blocks can 
be observed in the scheme of Figure~\ref{fig:architecture}. We remark that the network is fully siamese: all the parameters of the convolutional, 
global pooling and inner product layers are shared between the two branches. Importantly, the front-end (base architecture) can be 
arbitrarily chosen as long as it is fully convolutional. In the experimental section we show results with three different base architectures.

We chose to use a siamese network in order to be as close as possible to the philosophy of previous studies on visual-based virality 
prediction~\cite{deza2015understanding}. More formally, we assume the existence of a training set ${\cal T}$ consisting of $M$ pairs of images and 
the annotated relative virality: ${\cal T} = \{(I_m,J_m,v_m)\}_{m=1}^M$, where $v_m=1$ if $I_m$ is more viral than $J_m$ and $v_m=-1$ otherwise. In 
order to train the parameters of the siamese network, we optimize the sigmoid cross-entropy loss over the training set using stochastic gradient 
descent:
\begin{equation}
 {\cal L} = \sum_{m=1}^M -v_m \log \hat{v}_m - (1-v_m)\log (1-\hat{v}_m)),\label{eq:loss}
\end{equation}
where $\hat{v}_m = s(I_m;\theta)-s(J_m;\theta)$ is the subtraction of the output of the two branches of the siamese network, and $\theta$ denotes the 
set of shared parameters.

By designing the network as in Figure~\ref{fig:architecture}, the convolutional front-end will extract a set of feature maps, also known 
as latent concept detectors~\cite{zhou2015learning,porzi2015predicting}. While in previous studies these concepts were associated to the 
presence/absence of objects in the image or to the safety of urban scenes, in our case the latent detectors will be associated to virality. In this 
paper, we adopt global pooling so as to exploit these latent detectors for virality prediction and weakly-supervised virality localization.

\subsection{Global pooling}
We assume the existence of $L$ latent concept detectors and attempt to learn their relationship with virality. Each of these concept detectors is one 
output channel of size $W\times H$ of the convolutional front-end, $f_l\in\mathbb{R}^{W\times H}$ (each of the colored slices of \texttt{Conv N} in 
Figure~\ref{fig:architecture}). Generally speaking, global pooling extracts 
activations from each latent detector and feeds them to a fully connected layer responsible for classification. We remark the existence of three 
global pooling strategies in the literature.\vspace{-3mm}

\paragraph{Global average pooling} In~\cite{zhou2015learning}, the features maps are channel-wise averaged and fed to the fully connected layer. The 
classification score for each class $k\in\{1,\ldots,K\}$ (before the soft-max operation) is given by:
\begin{equation}
 q^{\textsc{gap}}_k = \sum_{l=1}^L w^{\textsc{gap}}_{kl} \frac{1}{WH}\sum_{w,h=1}^{W,H} f_l(w,h),
\end{equation}
were $w^{\textsc{gap}}_{kl}$ are the weights of the classification layer. This strategy is referred to as global average pooling (GAP) since all the 
pixels of channel $l$ are averaged before being fed to the fully 
connected layer. One prominent advantage of global pooling is that we can easily construct a \textit{class activation map} for each class 
$k$, which in case of GAP writes:
\begin{equation}
 a^{\textsc{gap}}_{k}(w,h) = \sum_{l=1}^L w^{\textsc{gap}}_{kl} f_l(w,h).\label{eq:act_gap}
\end{equation}
\vspace{-7mm}

\paragraph{Global max pooling} In case of global max pooling (GMP, see~\cite{oquab2015object} for details), the classification score writes:
\begin{equation}
 q^{\textsc{gmp}}_k = \sum_{l=1}^L w^{\textsc{gmp}}_{kl} \max_{w,h} f_l(w,h).
\end{equation}
And the class activation map associated to GMP is:
\begin{equation}
 a^{\textsc{gmp}}_{k}(w,h) = \sum_{l=1}^L w^{\textsc{gmp}}_{kl} f^{0}_l(w,h),\label{eq:act_gmp}
\end{equation}
where:\footnote{The choice of the notation $f^{0}_l$ will become clear later on.}
\begin{equation*}
 f^{0}_l(w,h) = \left\{\begin{array}{ll}
f(w,h) & \text{if } (w,h)=\displaystyle\arg\max_{w',h'}f_l(w',h'),\\
0 & \text{otherwise.}
\end{array}\right.\vspace{-3mm}
\end{equation*}

\paragraph{Global top-$N$ average pooling}
Intuitively, while average pooling takes all the pixels into account, max pooling takes only one value into account. In between, global top-$N$ 
average pooling~\cite{porzi2015predicting} (GNAP) takes the average of the $N$ largest values in the feature map. Thus, both average and max pooling 
can be 
seen as particular cases of top-$N$ average pooling with $N=WH$ and $N=1$ respectively. More formally, if $\eta\in[0,1]$ defines the proportion of 
pixels in the feature map to be averaged, we set $N_\eta = 1 + \lceil \eta\,(WH-1) \rceil$, so that $N_0=1$ and $N_1=WH$. With this notation, the 
top-$N_\eta$ average pooling writes:
\begin{equation}
 q^{\textsc{gnap}}_k = \sum_{l=1}^L w^{\textsc{gnap}}_{kl} \frac{1}{N_{\eta_l}}\sum_{(w,h)\in{\cal N}_l^{\eta_l}} f_l(w,h),
 \label{eq:nap}
\end{equation}
where ${\cal N}_l^{\eta_l}$ is the set of indices corresponding to the largest $N_{\eta_l}$ values of $f_l$. The associated class activation maps 
are:
\begin{equation}
 a^{\textsc{gnap}}_{k}(w,h) = \sum_{l=1}^L w^{\textsc{gnap}}_{kl} f^{\eta_l}_l(w,h),\label{eq:act_glenap}
\end{equation}
with:
\begin{equation*}
 f^{\eta_l}_l(w,h) = \left\{\begin{array}{ll}
f(w,h) & \text{if } (w,h)\in{\cal N}^{\eta_l}_l,\\
0 & \text{otherwise.}
\end{array}\right.
\end{equation*}
We now remark that the notation $f^0_l$ is justified since GNAP with $\eta=0$ corresponds to GMP. In addition, we note that GAP can be 
expressed as GNAP with $\eta=1$, and thus we can write: $a^{\textsc{gap}}_{k}(w,h) = \sum_{l=1}^L w^{\textsc{gap}}_{kl} f^1_l(w,h)$. 

Even if the top-$N$ average pooling may seem a good idea that generalizes the concept of average and max pooling, we are left with the tedious task 
of setting $\eta_l$. In order to avoid heuristics or the unaffordable process of cross-validation, we present an efficient way to estimate the 
gradient of the loss with respect to $\eta_l$, so that learning $\eta_l$ is included within the stochastic gradient descent optimization.

\subsection{Global learned top-$N$ average pooling}
Providing a formulation of the gradient with respect to $\eta_l$ requires understanding the behavior of the top-$N_\eta$ average with 
respect to $\eta$, since the dependence of $q^{\textsc{gnap}}_k$ with $\eta_l$ is not differentiable. In this section we propose a very efficient and 
intuitive way to approximate this gradient. Very importantly, the definition of the top-$N$ average pooling given above, and thus the 
formalization in this section, are general and independent of the applicative scenario.

We assume the back-propagation algorithm is able to compute the derivatives of the loss ${\cal L}$ with respect to $q_k^{\textsc{gnap}}$, so that 
we can use the chain rule to compute the derivative with respect to $\eta_l$ using:
\begin{equation}
 \frac{\partial {\cal L}}{\partial \eta_l} = \sum_{k=1}^K \frac{\partial {\cal L}}{\partial q_k^{\textsc{gnap}}} 
\frac{\partial q_k^{\textsc{gnap}}}{\partial \eta_l}.
\end{equation}
We only require now to give an expression for $\frac{\partial q_k^{\textsc{gnap}}}{\partial \eta_l}$. In order to do that, we first observe that, 
from~(\ref{eq:nap}) we have:
\begin{equation}
 \frac{\partial q_k^{\textsc{gnap}}}{\partial \eta_l} = \sum_{l=1}^L w^{\textsc{gnap}}_{kl} \frac{\partial g_l(\eta_l)}{\partial \eta_l},
\end{equation}
where we defined $g_l(\eta_l) =  \frac{1}{N_{\eta_l}}\sum_{(w,h)\in{\cal N}_l^{\eta_l}} f_l(w,h)$. 

\begin{figure}[t]
 \centering
 \includegraphics[width=0.85\columnwidth]{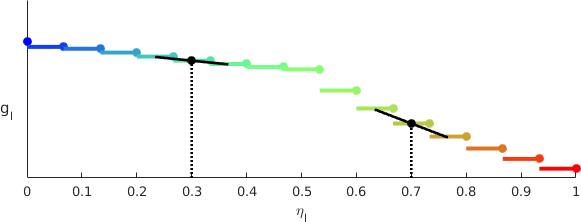}
 \caption{Example of the dependency of $g_l$ with $\eta_l$. Even if the original function is clearly non-differentiable, we can approximate the trend 
of the function very efficiently (black lines).\label{fig:gl_etal}}\vspace{-4mm}
\end{figure}

We need to compute the derivative of $g_l(\eta_l)$ with respect to $\eta_l$. Unfortunately, the function $g_l(\eta_l)$ is not 
differentiable with respect to $\eta_l$ everywhere. Moreover, at the points where the derivative is well-defined, it does not describe the trend of 
$g_l$. Indeed, the derivative is undefined at integer multiples of $\delta = (WH-1)^{-1}$ and zero elsewhere. Figure~\ref{fig:gl_etal} shows an 
example of $g_l(\eta_l)$ as a function of $\eta_l$. In all, we opt to ignore the exact derivative (when available) and try to understand 
the trend of the function $g_l$ instead. 

We adopt a very intuitive strategy that leads to an efficient implementation: approximate $g_l(\eta_l)$ by a second degree 
polynomial (parabola) and use the derivative of the latter as a proxy for the trend of the original function. 
One could think that fitting $L$ parabolas (one per channel) at every backward pass of the network is computationally intensive, since the 
coefficients of these parabolas need to be computed. However, if we carefully choose the fitting points of the parabola, the estimate of the 
derivative comes almost for free. Indeed, if $\eta_l^0$ denotes the current value of $\eta_l$, we fit the parabola at abscissae 
$\eta_l^0-\delta,\eta_l^0,\eta_l^0+\delta$, because the corresponding values $g_l(\eta_l^0-\delta),g_l(\eta_l^0),g_l(\eta_l^0+\delta)$ are the 
top-$N_{\eta_l^0}-1,N_{\eta_l^0},N_{\eta_l^0}+1$ averages respectively. Moreover, the derivative of such fit parabola at $\eta_l^0$ writes:
\begin{equation}
 \left.\frac{\partial g_l(\eta_l)}{\partial \eta_l}\right|_{\eta_l^0} = \frac{g_l(\eta_l^0+\delta)-g_l(\eta_l^0-\delta)}{2\delta}\label{eq:der-app}.
\end{equation}

Very importantly this strategy comes at almost no computational cost when compared to performing only the forward pass. This is because the most 
computationally intense operation is sorting the pixels of the feature map (this is required by the forward pass anyway).\footnote{Our CPU 
implementation in the worst case (when fine-tunning only LENA) increases the iteration time by $1.5$\% compared to GAP/GMP.} Once 
this is done, we need 
to compute the top-$N_{\eta_l^0}-1$ average, and update it to obtain the top-$N_{\eta_l^0}$ average and the top-$N_{\eta_l^0}+1$ average, but the 
overall computational cost is an average of $N_{\eta_l^0}+1$ real numbers. While the top-$N_{\eta_l^0}$ is used for the forward pass, the other two 
averages are used to estimate the trend of $g(\eta_l)$ using~(\ref{eq:der-app}), to further update the value of $\eta_l$.

When back-propagating down to the layer below, the memory requirements of the LENA layer are slightly higher than max pooling or average pooling. 
This is because this layer requires to store the $N_{\eta_l^0}$ pixel positions of the feature map that contributed to the forward pass, so that the 
layer propagates the error only to these pixels. Formally:
\begin{equation}
 \frac{\partial g_l(\eta_l)}{\partial f_l(w,h)} = \left\{\begin{array}{ll}
(N_l^{\eta_l^0})^{-1} & \text{if } (w,h)\in{\cal N}^{\eta_l^0}_l,\\
0 & \text{otherwise.}
\end{array}\right.
\end{equation}

We expect the LENA layer to be able to learn from the data which channels need to go through average pooling, which ones through max pooling and 
which ones require an intermediate option. Before describing the experimental protocol and showing the results, we briefly discuss how do we 
include objectness maps in our viraliency framework.

\subsection{Incorporating objectness}
Intuitively, virality is related to the presence of objects in the images and in order to ascertain the veracity of this statement, we also 
devise a straightforward strategy to include objectness information in our formulation. We choose to use objectness maps that in our case correspond 
to the class activation maps of~\cite{zhou2015learning} and are computed in the following way. First, we classify all the training images with 
AlexNet pretrained on ImageNet to extract the 30 most activated classes in the datasets we use.
We then generate the class activations maps of these classes for each image of the test and training sets. The objectness maps are then 
concatenated\footnote{A bilinear filter implemented as a deconvolutional layer was used to resize the objectness maps into the size of the 
feature maps, if needed.} to the feature maps of the siamese network (right before the global pooling, hence to 
\texttt{Conv N} in Figure~\ref{fig:architecture}). An extra convolutional layer is used to fuse the objectness maps with the latent concepts, and 
produce the same number of feature maps, that now include objectness information.
\vspace{-2mm}
%
%

\section{Experimental validation}
\subsection{Datasets and experimental protocol}
In order to assess the effectiveness of the proposed approach for virality prediction and localization, we perform experiments 
on two recently published datasets: the 
understanding image virality (UIV) dataset~\cite{deza2015understanding} and the image virality on GooglePlus (IVGP) 
dataset~\cite{guerini2013exploring}. In the following, we describe the experimental protocol, including datasets, network architectures and 
baselines.\vspace{-3mm}

\begin{figure}
\newlength{\mywidth}
\setlength{\mywidth}{1.4cm}
\hspace{-4mm}
 \begin{tabular}{cc}
  \hspace{1mm}UIV & IVPG \\
  \includegraphics[width=\mywidth,height=\mywidth]{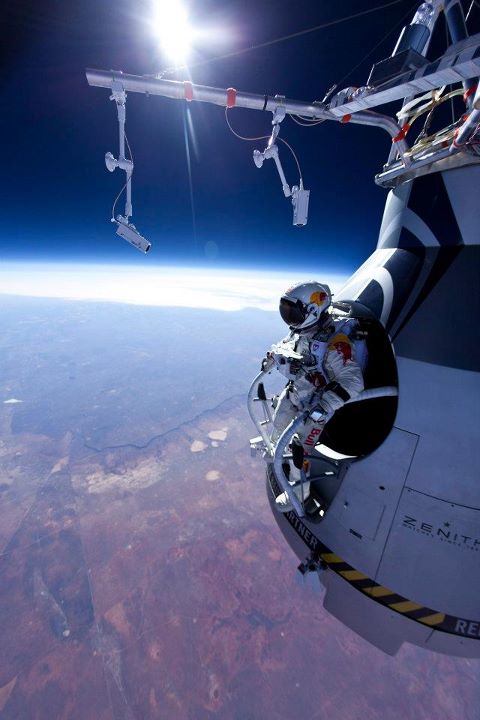}%
  \includegraphics[width=\mywidth,height=\mywidth]{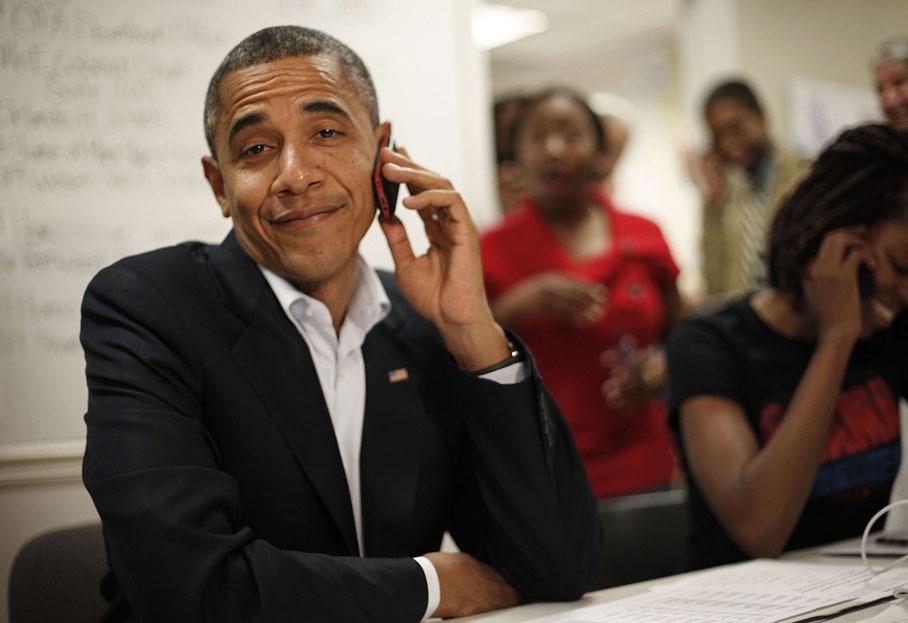}%
  \includegraphics[width=\mywidth,height=\mywidth]{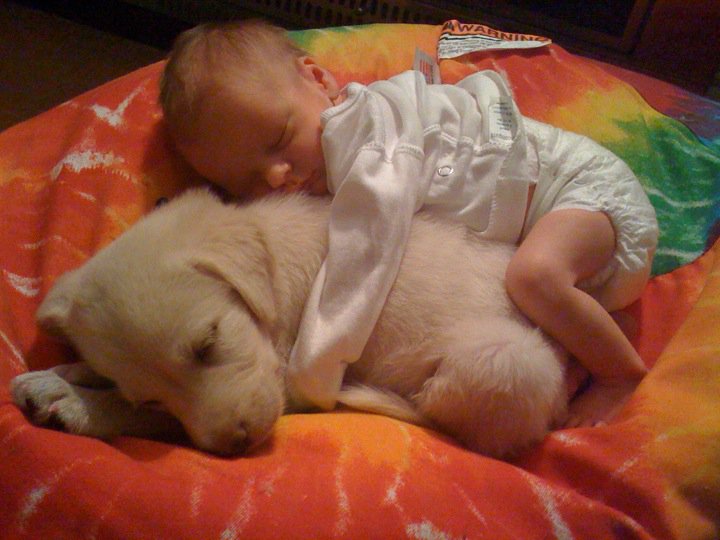}\hspace{-2mm}
  &
  \includegraphics[width=\mywidth,height=\mywidth]{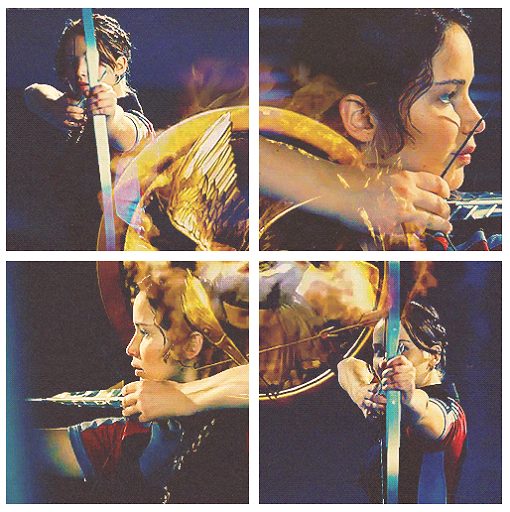}%
  \includegraphics[width=\mywidth,height=\mywidth]{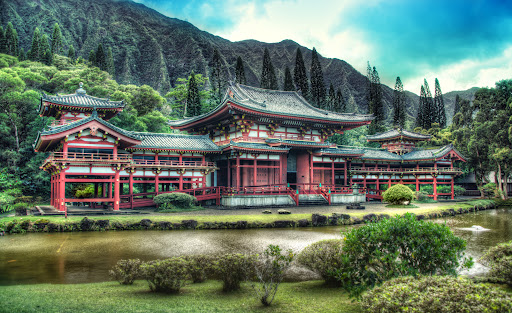}%
  \includegraphics[width=\mywidth,height=\mywidth]{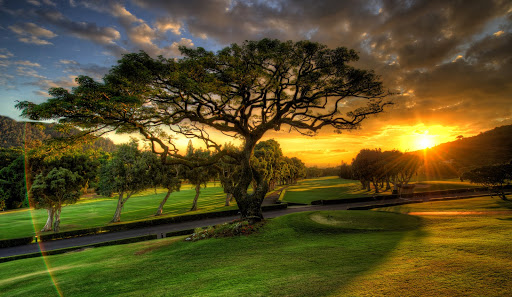}\vspace{-1mm}\\
  \includegraphics[width=\mywidth,height=\mywidth]{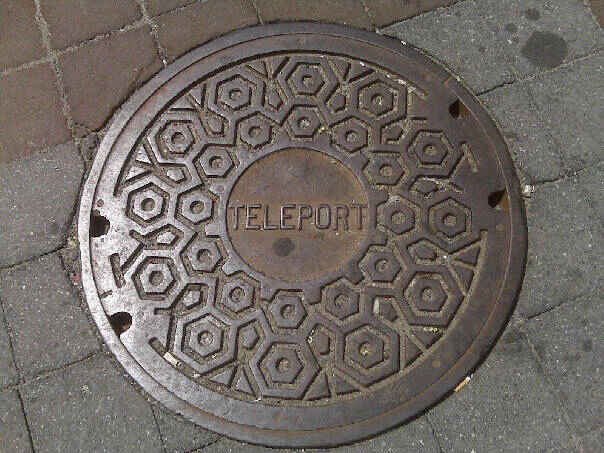}%
  \includegraphics[width=\mywidth,height=\mywidth]{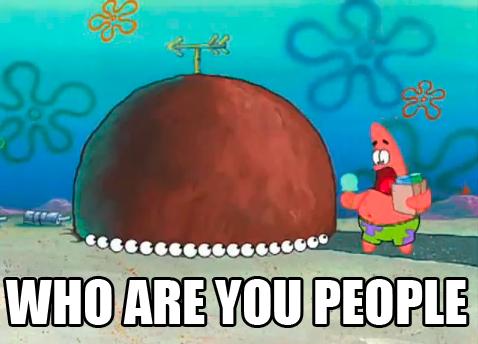}%
  \includegraphics[width=\mywidth,height=\mywidth]{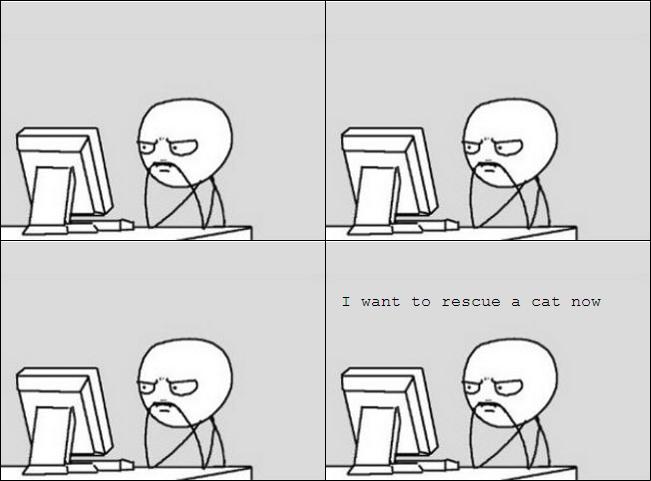}\hspace{-2mm}
  &
  \includegraphics[width=\mywidth,height=\mywidth]{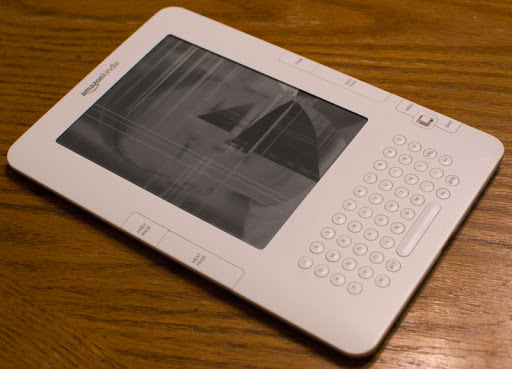}%
  \includegraphics[width=\mywidth,height=\mywidth]{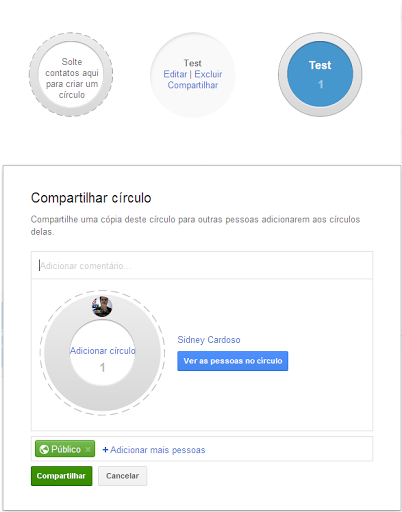}%
  \includegraphics[width=\mywidth,height=\mywidth]{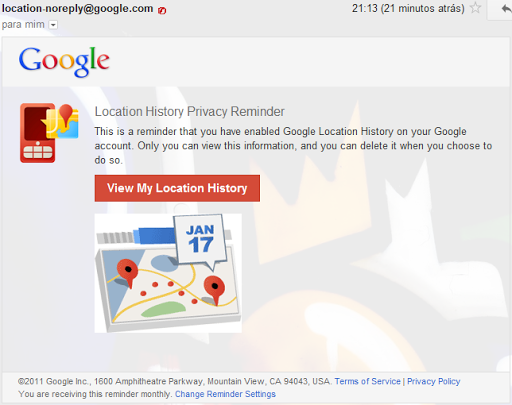}\\
 \end{tabular}
 \caption{Sample most (top) and least (bottom) viral images from the UIV (left) and the IVGP (right) datasets.\label{fig:sample}}\vspace{-6mm}
\end{figure}

\paragraph{Datasets.} The \textbf{UIV} dataset~\cite{deza2015understanding} consists on $10$K+ Reddit images with the associated virality score (a 
detailed explanation of the dataset construction can be found in the original paper). In~\cite{deza2015understanding} the data are organized in 
pairs, such as to predict relative virality scores, and a training and a test set of $4,\!550$ and $489$ images pairs are created. Since the 
insufficiency of data can easily lead to overfitting problems when training deep architectures, we created a much larger dataset for our experiments.
Inspired by how the training and test sets are generated in~\cite{deza2015understanding}, we randomly created a test set of 
$2,\!965$ image pairs, taking one sample from the $250$ most viral images and one from the $250$ least viral images.  
The training set consists on $18,\!182$ randomly generated image pairs, containing one image with above-median virality and one image with 
below-media virality. Very importantly, we ensured that the training and test sets are disjoint, so that the test pairs are not used during 
training.

The \textbf{IVGP} dataset presented in~\cite{guerini2013exploring} consists on GooglePlus images of the top followed profiles in this social 
network. Images were gathered from the most followed profiles to avoid ``friendship dynamics'' when reposting content and to ensure enough 
visibility to each image (see~\cite{guerini2013exploring} for more details). Intuitively this guarantees that all images go through a minimum number 
of impartial views, and therefore the measures of virality are significant. The original dataset has $150$K+ images, but only 
$90$K are currently available.\footnote{Image are available only through the users' public profile.} After assessing their virality with the 
formulation in~\cite{deza2015understanding}, we generated $11,\!704$ and $2,\!926$ image pairs for training and test respectively. 
Each image pair consists of one of the $15$K most viral images and one of the $15$~K least viral images. The training and test sets are disjoint. 

Sample images of both datasets are shown in Figure~\ref{fig:sample}.\vspace{-4mm}

\paragraph{Network architectures.} We used three different base architectures for the proposed siamese network. Specifically, we consider
the five convolutional layers of AlexNet \cite{krizhevsky2012imagenet} (Alex5), \new{similarly to~\cite{zhou2015learning} we append two convolutional 
layers with 512 units, \texttt{3x3} kernel and stride 1 to Alex5 leading to 7 convolutional layers (Alex7)\footnote{Both the architecture and 
the trained model used to initialize it are publicly available in \url{https://github.com/metalbubble/CAM}.}}
and finally the 13 convolutional layers of the VGG16 network~\cite{simonyan2014very} (VGG13). All the networks have been fine-tuned 
for $10$K iterations and the learning rate policy was fixed for all experiments using the same base architecture. The weights of these 
networks were initialized from pretrained ImageNet models. 
The training protocol details can be found in the supplementary material. The code of the LENA layer and all the trained models will be made freely 
accessible for research purposes.\vspace{-4mm}

\paragraph{Baselines.} We compare the proposed method with several baselines. Deza \& Parikh \cite{deza2015understanding}
is the only previous approach addressing virality prediction and considers an SVM  
with features extracted from the sixth layer of the AlexNet network. Importantly, we could not use visual attributes because they are available 
only for a small subset of the dataset in~\cite{deza2015understanding} (and not for~\cite{guerini2013exploring}) and in addition it is not 
straightforward to extract them in an automatic manner. In order to evaluate the effectiveness of the proposed LENA layer, we 
compare it with global max pooling (GMP)~\cite{oquab2015object}, global average pooling (GAP)~\cite{zhou2015learning} and with top-$N$ average 
pooling (GNAP)~\cite{porzi2015predicting}. Regarding the LENA layer, we try different initializations for $\eta_l$, namely $0$, $1/2$ and $1$, and 
denote them by G\textsc{lena}P$-0$, $-$\textonehalf~and $-1$.

\begin{table}[t]
 \caption{Accuracy results on predicting virality with Alex7 on the UIV and IVGP datasets. The two last columns, 
$\rightarrow$UIV and $\rightarrow$IVPG correspond to cross-dataset results, training in IVPG and testing in UIV and viceversa, 
respectively.\vspace{0.1cm}\label{tab:virality_prediction}}
\centering
\scalebox{0.7}{%
 \begin{tabular}{cccccc}
 \toprule
  Method     & UIV & IVGP && $\rightarrow$UIV &\hspace{-4mm} $\rightarrow$IVPG\\
 \midrule
  Deza \& Parikh~\cite{deza2015understanding} & 59.5 & 65.4 && 51.4 &\hspace{-4mm} 48.5 \\
 \midrule
  GAP~\cite{zhou2015learning}          & 61.4 & 68.0 && 54.0 &\hspace{-4mm} 52.0 \\
  GMP~\cite{oquab2015object}          & 62.2 & 71.0 && 56.2 &\hspace{-4mm} 52.3\\
  GNAP~\cite{porzi2015predicting}      & 62.3 & 71.2 && {\bf 57.3} &\hspace{-4mm} {\bf 52.7} \\ 
 \midrule
  G\textsc{lena}P$-$0                         & {\bf 62.7} & 71.3 && 55.9 &\hspace{-4mm} 52.0\\
  G\textsc{lena}P$-$\textonehalf              & 61.5 & 71.6 && 57.1 &\hspace{-4mm} {\bf 52.7}\\
  G\textsc{lena}P$-$1                         & 62.6 & {\bf 72.7} && 55.9 &\hspace{-4mm} 52.3\\
 \bottomrule
 \end{tabular}}\vspace{-4mm}
\end{table}

\subsection{Predicting virality}

We first evaluate the performance on virality prediction. Table~\ref{tab:virality_prediction} shows the accuracy of the different methods on the UIV 
and IVPG datasets. The first two columns correspond to the standard training and testing, while the third and fourth columns 
to cross-dataset experiments. For instance, $\rightarrow$UIV means training on IVPG and testing on UIV. 

We first observe that all the end-to-end trainable models systematically outperform the SVM-based method in \cite{deza2015understanding},\footnote{We attribute the small 
improvement of the baseline over what was reported in~\cite{deza2015understanding} to our larger dataset.} which is in agreement to the 
findings of the community in a wide variety of vision applications. Also, in agreement with previous studies~\cite{zhou2015learning}, we found 
that embedding a global pooling layer into a specific architecture (\eg AlexNet)
is outperformed by considering a corresponding fully connected layer within the same network (by 1.5 and by 1.7 points on UIV and IVGP, 
respectively, numbers not reported in the table). We remark that this slight increase of performance comes at the cost of completely losing the 
ability to localize the viral parts of the image (as also discussed in~\cite{zhou2015learning} for weakly-supervised object detection). Thirdly, for 
the ``within dataset'' experiments (training and test belong to the same dataset), we remark that at least one of the initializations of the proposed 
LENA pooling is systematically outperforming all the 
baseline methods. Regarding the cross-dataset experiments, we highlight the inability of all the methods to maintain the same virality recognition 
performance. Finally, when comparing the performance dataset-wise we realize that: (i) the 
accuracy on the within dataset experiments for IVPG are higher than for UIV and (ii) more importantly, the performance on $\rightarrow$UIV are 
systematically better than for the $\rightarrow$IVPG experiments. This would suggest that the IVPG contains data allowing better generalization than 
UIV. For the rest of the present study, we \new{intensively exploit} the IVPG dataset.

\subsection{The use of objectness maps}
Naturally, one may wonder if prior knowledge of which objects are in the image (and where are them) could help predicting virality. In order to 
analyse this aspect, we performed experiments that take the objectness of the images into account. Table~\ref{tab:objectness} 
reports the accuracy results on virality prediction with the three base architectures (Alex5, Alex7 and VGG13) on the IVGP dataset with (w/) and 
without (w/o) objectness (the third column of Table~\ref{tab:objectness} corresponds to the second column of 
Table~\ref{tab:virality_prediction}).

We first observe that the use of objectness is advantageous: in most of the cases the accuracy raises when adding objectness. Second, we 
notice that VGG13 results are systematically higher than Alex7 ones, 
independently of the objectness. In other words, for a given method the minimum over the fifth and sixth columns is always higher than the maximum 
over the third and fourth columns. A similar trend is found when comparing the performance of the Alex7 and Alex5 networks. Finally, we highlight 
that, as in the case of Table~\ref{tab:virality_prediction}, the best accuracy across the initializations of LENA is consistently superior to the 
three baselines, independently of the base architecture and of the use of objectness maps (except for Alex5 with objectness). This reinforces the 
idea that learning $\eta$ within a global pooling scheme at the top of a convolutional network helps the virality prediction task.

\begin{table}
 \caption{Virality prediction accuracy for the three base architectures (Alex5, Alex7 and VGG13) with (w/) and without (w/o) 
objectness on the IVGP dataset.\label{tab:objectness}\vspace{0.1cm}}
 \centering
 \scalebox{0.6}{%
 \begin{tabular}{cccccccccccc}
 \toprule
  \multirow{2}{*}{Method}    & \multicolumn{2}{c}{Alex5} & \multicolumn{2}{c}{Alex7} & \multicolumn{2}{c}{VGG13} & \multicolumn{2}{c}{VGG16} & 
\multicolumn{3}{c}{Res50}\\
	    &  w/o & w/ &  w/o & w/ &  w/o & w/ &  FT-345 & FT-all &  5c & All-5 & FT-45\\
 \midrule
  GAP~\cite{zhou2015learning}       & 68.0 & {\bf 71.2} & 68.0 & 71.1 & 71.1 & 74.1 & 70.6 &  & 73.5 &  & 74.5 \\
  GMP~\cite{oquab2015object}        & 68.4 & 70.7 & 71.0 & 71.6 & 74.1 & 73.2 & 75.0 & 72.8 & 70.1 & \\
  GNAP~\cite{porzi2015predicting}   & 68.6 & 70.3 & 71.2 & 71.6 & 72.4 & 74.9 & 72.6 & & -- & \\ 
 \midrule
  G\textsc{lena}P$-$0               & {\bf 69.7} & 70.2 & 71.3 & {\bf 72.6} & 73.4 & {\bf 75.6} & 73.0 & & 70.1 &\\
  G\textsc{lena}P$-$\textonehalf    & 67.9 & 69.8 & 71.6 & 72.2 & 73.4 & 74.0 & 73.7 & & 70.7 &\\
  G\textsc{lena}P$-$1               & 66.9 & 69.1 & {\bf 72.7} & {\bf 72.6} & {\bf 75.7} & 74.7 & 75.2 & 74.7 & -- & 73.4 & 74.5 \\
 \bottomrule
 \end{tabular}%
 }\vspace{-4mm}
\end{table}

\subsection{Viraliency maps}

\begin{figure*}
\setlength{\mywidth}{1.8cm}
\hspace{-0.2cm}%
\centering
 \begin{tabular}{cccccccc}
 Original & \multicolumn{3}{c}{No objectness} && \multicolumn{3}{c}{Objectness}\\
 \cmidrule{2-8}
 Image & GAP~\cite{zhou2015learning} & GMP~\cite{oquab2015object} & G\textsc{lena}P && GAP~\cite{zhou2015learning} & GMP~\cite{oquab2015object} & 
G\textsc{lena}P \\
\includegraphics[width=\mywidth,height=\mywidth]{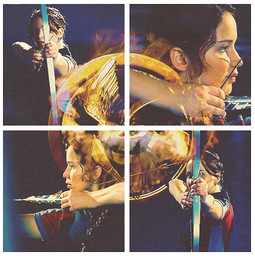}&%
\includegraphics[width=\mywidth,height=\mywidth]{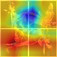}&\hspace{-2mm}%
\includegraphics[width=\mywidth,height=\mywidth]{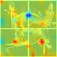}&\hspace{-2mm}%
\includegraphics[width=\mywidth,height=\mywidth]{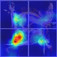}&&\hspace{-2mm}%
\includegraphics[width=\mywidth,height=\mywidth]{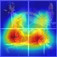}&\hspace{-2mm}%
\includegraphics[width=\mywidth,height=\mywidth]{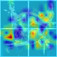}&\hspace{-2mm}%
\includegraphics[width=\mywidth,height=\mywidth]{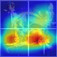}\vspace{-1.5mm}\\
\includegraphics[width=\mywidth,height=\mywidth]{img/sample_images/ivgp-top_b.jpg}&%
\includegraphics[width=\mywidth,height=\mywidth]{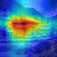}&\hspace{-2mm}%
\includegraphics[width=\mywidth,height=\mywidth]{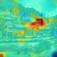}&\hspace{-2mm}%
\includegraphics[width=\mywidth,height=\mywidth]{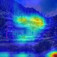}&&\hspace{-2mm}%
\includegraphics[width=\mywidth,height=\mywidth]{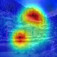}&\hspace{-2mm}%
\includegraphics[width=\mywidth,height=\mywidth]{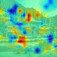}&\hspace{-2mm}%
\includegraphics[width=\mywidth,height=\mywidth]{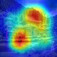}\vspace{-1.5mm}\\
\includegraphics[width=\mywidth,height=\mywidth]{img/sample_images/ivgp-top_c.jpg}&%
\includegraphics[width=\mywidth,height=\mywidth]{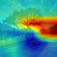}&\hspace{-2mm}%
\includegraphics[width=\mywidth,height=\mywidth]{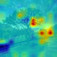}&\hspace{-2mm}%
\includegraphics[width=\mywidth,height=\mywidth]{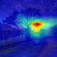}&&\hspace{-2mm}%
\includegraphics[width=\mywidth,height=\mywidth]{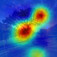}&\hspace{-2mm}%
\includegraphics[width=\mywidth,height=\mywidth]{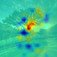}&\hspace{-2mm}%
\includegraphics[width=\mywidth,height=\mywidth]{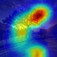}\\
 \end{tabular}
 \caption{Viraliency (class activation) maps for three images of the IVGP dataset using the Alex7 base network. The four 
columns correspond to (left 
to right) the original image, viraliency for GAP, for GMP and for G\textsc{lena}P$-$1, without and with 
objectness.\label{fig:viraliency}}\vspace{-3mm}
\end{figure*}

\begin{figure}[t]
 \setlength{\mywidth}{1.5cm}
\centering
 \begin{tabular}{ccccc}
\includegraphics[width=\mywidth,height=\mywidth]{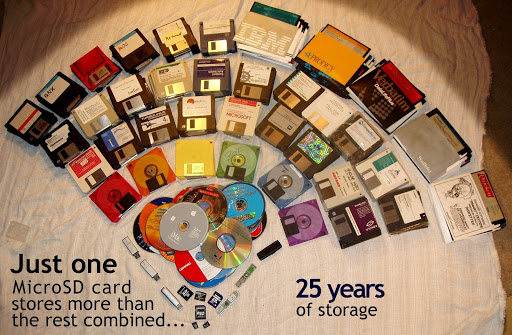}&\hspace{-4mm}%
\includegraphics[width=\mywidth,height=\mywidth]{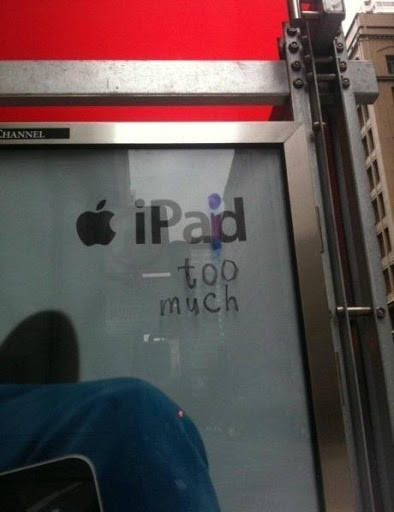}&\hspace{-4mm}%
\includegraphics[width=\mywidth,height=\mywidth]{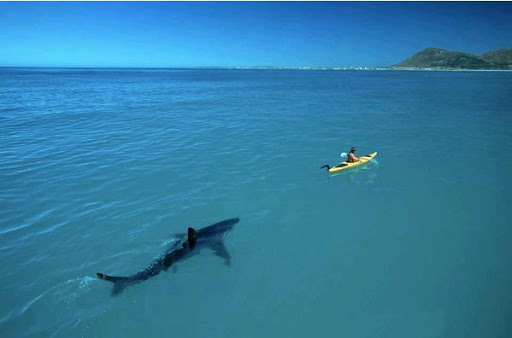}&\hspace{-4mm}%
\includegraphics[width=\mywidth,height=\mywidth]{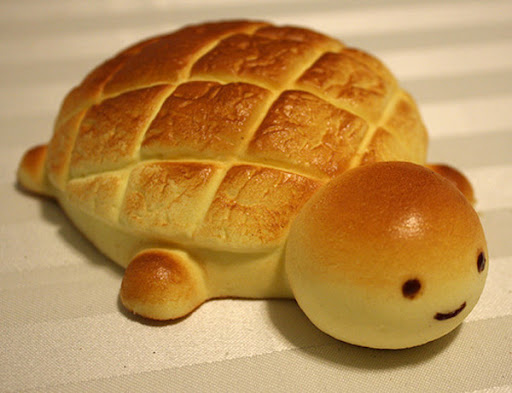}&\hspace{-4mm}%
\includegraphics[width=\mywidth,height=\mywidth]{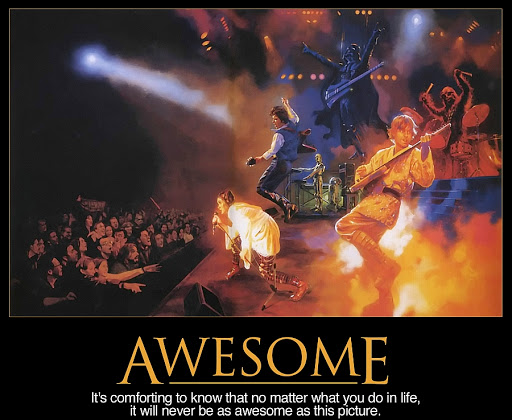}\vspace{-2mm}\\
\includegraphics[width=\mywidth,height=\mywidth]{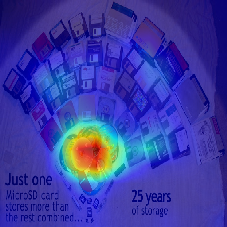}&\hspace{-4mm}%
\includegraphics[width=\mywidth,height=\mywidth]{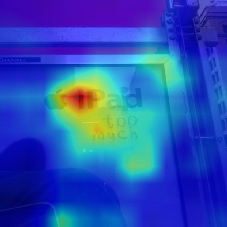}&\hspace{-4mm}%
\includegraphics[width=\mywidth,height=\mywidth]{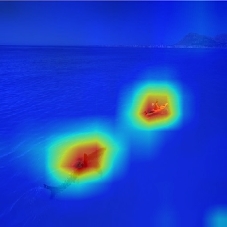}&\hspace{-4mm}%
\includegraphics[width=\mywidth,height=\mywidth]{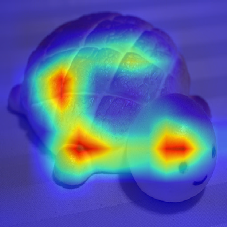}&\hspace{-4mm}%
\includegraphics[width=\mywidth,height=\mywidth]{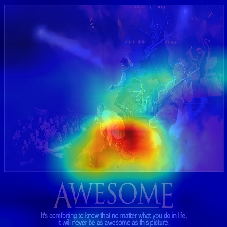}\\
 \end{tabular}
 \caption{Viraliency maps from G\textsc{lena}P of five images of IVGP.\label{fig:extravmaps}}\vspace{-7mm}
\end{figure}

One of the prominent features of the global pooling layers is their capacity to implicitly localize the objects and 
concepts through the 
analysis of the class activation maps~\cite{oquab2015object,zhou2015learning}. More importantly, this is achieved with 
weak supervision: no 
localization information is used during training. In the precise case of virality prediction these maps correspond to 
the virally salient
parts of the image, \ie the viraliency maps. Figure~\ref{fig:viraliency} plots the viraliency maps superposed to three 
of the top viral images 
of the IVGP dataset for GAP~(\ref{eq:act_gap}), GMP~(\ref{eq:act_gmp}) and G\textsc{lena}P$-$1~(\ref{eq:act_glenap}) 
(denoted by G\textsc{lena}P from 
now on), without obectness in the first three columns, and including objectness in the last three columns. When no 
objectness is used, the viraliency 
maps associated to the three pooling layers have clear distinct behaviors. Indeed, GAP seems to be able to point to a 
fairly compact zone of the 
image responsible for virality, while GMP highlights several small zones, thus leading to a viraliency map that is 
spread over the image. The proposed 
LENA pooling is able to spot a few (2 to 4) zones in the image responsible for virality. The use of objectness seems to 
bring the three global 
pooling layers towards a more similar behavior, as expected. Indeed, we can see that the viraliency maps of GAP and 
G\textsc{lena}P are now close to 
each other. Regarding GMP, even if the spread behavior observed when no objectness was used is still dominant, the more 
viral zones are aligned with 
the big bulbs in GAP and G\textsc{lena}P when objectness is used. Interestingly, we can observe in the examples that the 
use of objectness is a 
two-edged sword. For instance, the viraliency map of the second image with G\textsc{lena}P without objectness contains a 
very hot spot on the bottom 
right corner of the image, which disappears when using objectnes. At the same time, a bulb in the lower part of the 
third viraliency map for 
G\textsc{lena}P appears when objectness is included, but the main spot is widen to include the tree. These results 
confirm our initial hypothesis that 
richer global pooling strategies are adequate when recognizing subjective/abstract properties of images. More pictures 
of viraliency maps for all 
pooling strategies are shown in the supplementary material.

Figure~\ref{fig:extravmaps} shows some viraliency maps for the LENA pooling that are worth to be discussed. First of 
all, we observe that most of 
these viraliency maps consist on different virally salient areas, thanks to the ability of the proposed layer to pool 
information from several 
pixels. The capacity of the network to highlight the viral parts of an image using only visual information is 
remarkable: the network is 
\new{partially learning the complexity} associated to virality. Two immediate explanations of this phenomenon are the 
potential bias towards 
text and objects, as suggested by the second and third images respectively. However, the text in the first and fourth 
images does not belong to the 
highlighted region and the viraliency map of the first and fourth images does not match with the objects' regions. The 
fourth image is of special 
interest, because the network suggests that virality arises from the combination of ``pastry'' and ``turtle''.

\begin{figure*}[t]
 \setlength{\mywidth}{1.6cm}
 \centering
 \begin{tabular}{c|c|cccccc}
 Original Image & GAP/GMP & \multicolumn{5}{c}{G\textsc{lena}P}\\\hspace{-4mm}
   \multirow{2}{*}[13mm]{\includegraphics[width=2\mywidth,height=2\mywidth]{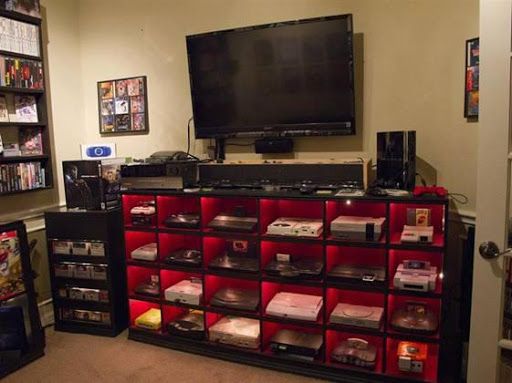}}&%
  \includegraphics[width=\mywidth]{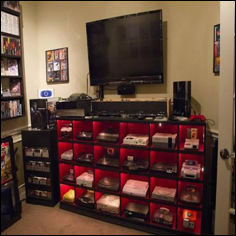}&%
  \includegraphics[width=\mywidth]{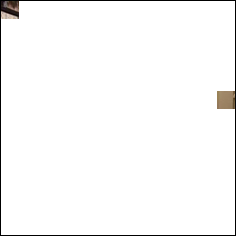}&\hspace{-3mm}%
  \includegraphics[width=\mywidth]{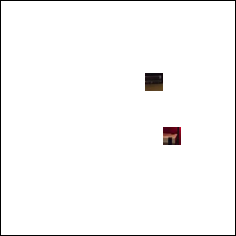}&\hspace{-3mm}%
  \includegraphics[width=\mywidth]{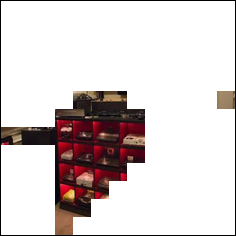}&\hspace{-3mm}%
  \includegraphics[width=\mywidth]{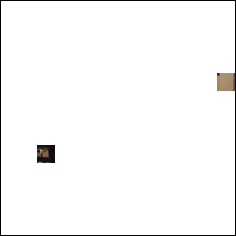}&\hspace{-3mm}%
  \includegraphics[width=\mywidth]{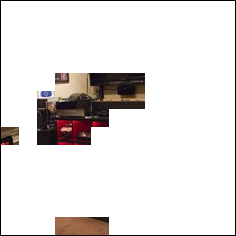}\vspace{0.5mm}\\%
  &%
  \includegraphics[width=\mywidth]{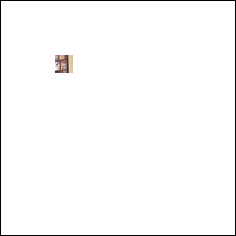}&%
  \includegraphics[width=\mywidth]{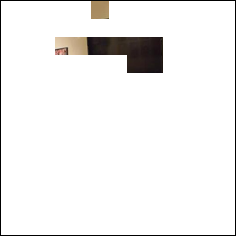}&\hspace{-3mm}%
  \includegraphics[width=\mywidth]{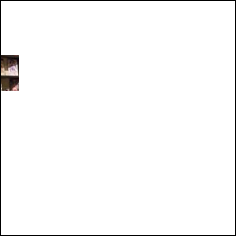}&\hspace{-3mm}%
  \includegraphics[width=\mywidth]{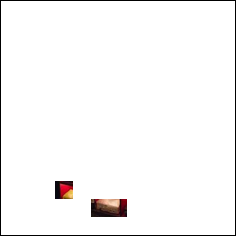}&\hspace{-3mm}%
  \includegraphics[width=\mywidth]{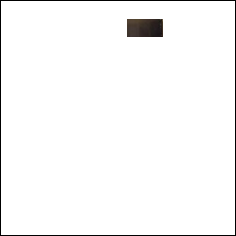}&\hspace{-3mm}%
  \includegraphics[width=\mywidth]{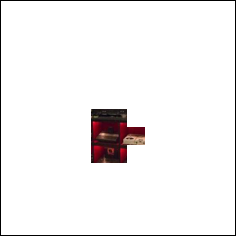}\\%
 \end{tabular}\vspace{1mm}
 \caption{Example of the receptive fields of different channels of the G\textsc{lena}P with Alex7 on an image from IVGP 
(on the left). The two 
most-left images correspond to the two extremes GAP (top) and GMP (bottom), where respectively all pixels and one pixel 
of the feature map are 
averaged. \new{The rest corresponds to receptive fields of different sizes (\ie $\eta_l$) learned by the LENA 
pooling layer.}\label{fig:contributors}}\vspace{-3mm}
\end{figure*}

\begin{table}
 \centering
\caption{Virality localization precision (Pr) and recall (Rc).\label{tab:local}\vspace{1mm}}
\scalebox{0.8}{
   \begin{tabular}{cccccccc}
   \toprule
 \multirow{2}{*}{Set} & \multirow{2}{*}{Meas.} & \multicolumn{3}{c}{No objectness} & \multicolumn{3}{c}{Objectness}\\
 \cmidrule{3-8}
 & & GAP & GMP & G\textsc{lena}P & GAP & GMP & G\textsc{lena}P \\
 \midrule
 \multirow{2}{*}{\rotatebox[origin=c]{90}{IVGP}}& Pr & 0.34 & 0.35 & 0.29 & 0.45 & 0.47 & 0.48 \\
 & Rc & 0.39 & 0.33 & 0.35 & 0.38 & 0.44 & 0.44 \\
 \cmidrule{2-8}
 \multirow{2}{*}{\rotatebox[origin=c]{90}{UIV}} & Pr & 0.37 & 0.43 & 0.41 & 0.41 & 0.47 & 0.45 \\
 & Rc & 0.33 & 0.31 & 0.35 & 0.36 & 0.31 & 0.38 \\
 \bottomrule
 \end{tabular}}\vspace{-6mm}
\end{table}

In order to provide a preliminarily quantitatively evaluation of the viraliency maps, we annotated the 150 most viral 
images of both datasets. 
The annotation process was carried out by a crew of 15 PhD students (9 male, 6 female). The task given to the annotators 
was: ``Given that this 
image is known to have gone viral, spot the part(s) of the image you believe they are responsible for this.'' These 
annotations will be released 
together with the code and the trained models. We consider a pixel-wise classification task, where the intensity of the 
viraliency map is taken as 
the probability of this pixel being viral, and report precision and recall in Table~\ref{tab:local}. Firstly, we remark 
that objectness helps the 
three pooling strategies, mostly by increasing their precision while not decreasing too much their recall. Second, GAP 
and G\textsc{lena}P outperform 
GMP by a significant margin. These results, combined with the prediction scores of Table~\ref{tab:objectness} suggest 
that G\textsc{lena}P is the 
most efficient layer for the simultaneous prediction and localization of virality.

To further push the understanding of the viraliency maps, Figure~\ref{fig:contributors} plots the receptive field of 
different output activations of 
the G\textsc{lena}P for an image of IVGP. The two most left correspond to channels averaging over all pixels (as GAP 
does) or over a single pixel (as 
GMP does), the rest correspond to channels with intermediate values of $\eta_l$. In terms of information propagation, 
the advantage of LENA is 
two-fold. On one hand many image regions can contribute to forward information to the classification layer, thus 
enlarging the forward capacity 
of max-pooling. On the other side the error is back-propagated only to those regions that contributed during the forward 
pass, leading to a more 
efficient strategy than average-pooling (that propagates the error everywhere). We believe that this low-level behavior 
explains the 
effectiveness exhibited by the LENA pooling.


\subsection{Sensitivity analysis}

\begin{figure}[t]
 \centering
 \includegraphics[width=.9\linewidth]{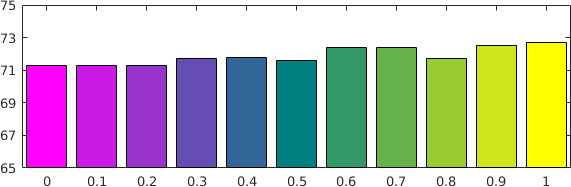}\\
 \caption{Sensitivity analysis of the accuracy of Alex7 on the IVGP dataset to different initialization values of 
$\eta$.\label{fig:sensitivity}}\vspace{-4mm}
\end{figure}

\begin{figure}[t]
 \centering
 \includegraphics[width=.9\linewidth]{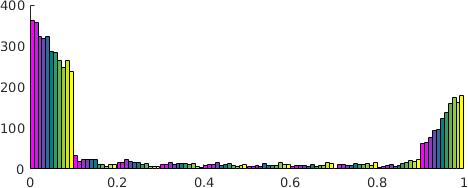}
  \caption{Histogram of the values of $\eta$ after convergence of the Alex7 network on the IVGP dataset for different values of initial $\eta$. The 
colors correspond to each of the initializations in Figure~\ref{fig:sensitivity}.\label{fig:histogram_eta}}\vspace{-5mm}
\end{figure}

We perform an analysis to study the sensitivity of the proposed LENA layer to the initial value of $\eta$. In details, we 
initialize the Alex7-based siamese structure with 11 different values of $\eta$ (from 0 to 1 every 0.1) and plot the accuracy of these different 
trainings in Figure~\ref{fig:sensitivity}. This confirms our intuition that the final accuracy does not exhibit strong dependencies to the initial 
value of $\eta$. To further analyze the potential behavioral differences of with respect to the initial values of $\eta$, we plot the histogram of 
converged values of $\eta$ in Figure~\ref{fig:histogram_eta}. The colors on this figure correspond to the colors on Figure~\ref{fig:sensitivity}.
Very importantly, we observe that independently of the initialization roughly one-third of the channels converge to average pooling and 
two-thirds to max pooling. This provides an explanation of why the G\textsc{lena}P strategy outperforms other global pooling strategies such as 
global average or max-pooling.

%

\section{Conclusions}
In this paper we addressed the task of simultaneous prediction and localization of virality using only visual information \new{and image-level 
annotations}. To this aim, we propose to 
use an end-to-end trainable siamese deep architecture with three main blocks: a convolutional front-end, a global pooling layer and a classification 
layer. Within this context, we introduced the LENA pooling layer, that estimates the optimal $\eta$ per each convolutional 
feature map. We performed an extensive experimental evaluation that shows the effectiveness of the proposed architecture, and of the LENA layer, for 
the simultaneous prediction and localization of virality. In the future we will assess the usefulness of such architectures for other subjective 
properties of visual data, as well as develop methods able to exploit other metadata related to virality, such as the comments in the social 
network associated to the image. Additionally, we plan to identify different temporal patterns of virality and design methods to recognize them in an 
automatic manner.

{\small
\bibliographystyle{ieee}
\bibliography{viraliency.bib}
}


\end{document}


\title{Supplementary material of the CVPR'17\\Viraliency: Pooling Local Virality\vspace{-5mm}}

\author{First Author\\
Institution1\\
Institution1 address\\
{\tt\small firstauthor@i1.org}
\and
Second Author\\
Institution2\\
First line of institution2 address\\
{\tt\small secondauthor@i2.org}
}

\newcommand{\todo}[1]{\textcolor{red}{#1}}

\maketitle

\section{Training details}
We implemented our LENA pooling layer within the Caffe framework and ran all our experiments using a Tesla K40 GPU. All the networks were fine-tuned 
from the convolutional filters obtained when training these networks for the $1,\!000$ image classification task on the ImageNet dataset. We iterated 
the stochastic gradient descent algorithm for $10,\!000$ iterations with a momentum of $\mu=0.9$ and a weight decay of $\lambda=0.05$. The learning 
rate followed a \texttt{step} policy with factor $0.1$ every $5,\!000$ iterations, with base learning rate set at $0.0001$.

\section{Virality Score}
While the virality scores for the UIV dataset were provided with the images, this information was not available for the IVGP dataset. In order to have 
similar measures of virality and therefore be able to compare results obtained on the two datasets, we adapted the annotation procedure 
of~\cite{deza2015understanding} to the metadata provided in~\cite{guerini2013exploring}. The original virality score was defined 
in~\cite{deza2015understanding} as:
\begin{equation}
	V_i = \frac{L_i}{\bar{L}} \log\left(\frac{M_i}{\bar{M}}\right),
\end{equation}
where $V_i$ is the virality score of the $i$-th image, $L_i$ and $\bar{L}$ are the number of likes associated to the $i$-th image and the 
average number of likes over the dataset, respectively, and $M_i$ and $\bar{M}$ are the number of resubmissions of image $i$ and the average number 
of resubmissions over the dataset, respectively. The terms ``likes'' and ``resubmissions'' are valid for images extracted from Reddit, but need to be 
adapted for 
images downloaded from GooglePlus. We intuitively choose to replace the ``resubmissions'' by ``reshares'' and ``likes'' by the difference of 
``upvotes'' minus ``downvotes'', and use the exact same formulation (previous equation) to retrieve the virality score of each image in IVGP.


%

\section{Extra viraliency maps}
In the next pages we present more examples of the viraliency maps obtained for different images with the three pooling strategies (GAP, GMP and 
G\textsc{lena}P) with and without objectness. We show results on the IVGP dataset from Figure~\ref{fig:ivgp-1} to Figure~\ref{fig:ivgp-5} and on the 
UIV dataset from Figure~\ref{fig:uiv-1} to Figure~\ref{fig:uiv-4}.

Generally speaking we can observe the following trends. First, GMP is generating spread viraliency maps, where many image locations are partly 
highlighted. Second, we observe that G\textsc{lena}P is able to produce viraliency maps composed of several strong active areas. Compared to GAP, 
this has 
the advantage of pointing to many virally salient locations in the images, since GAP is mostly producing one strong compact area. When 
objectness is 
added things may change in two directions depending on the image (and this holds for the three pooling strategies that tend to keep their relative 
differences). If the image contains clearly defined objects (\ie in the foreground), the viraliency maps tend to focus more on these objects. 
Otherwise, the viraliency maps are modified in a way that may be perceived as ``independent of the image content.'' This behavior is quite expected 
since objectness cannot provide strong localization cues if the objects in the image are not strongly localized.

We believe this large set of examples (chosen to show both the successes and failures of the proposed G\textsc{lena}P layer) provides rich insights 
on the capabilities of the novel layer for virality localization and on how does it function.

\begin{figure*}
\newlength{\mywidth}
\setlength{\mywidth}{2.3cm}
\centering
\begin{tabular}{cccccccc}
 Original & \multicolumn{3}{c}{No objectness} && \multicolumn{3}{c}{Objectness}\\
 \cmidrule{2-8}
 Image & GAP~\cite{zhou2015learning} & GMP~\cite{oquab2015object} & G\textsc{lena}P && GAP~\cite{zhou2015learning} & GMP~\cite{oquab2015object} & 
G\textsc{lena}P \\
\includegraphics[width=\mywidth,height=\mywidth]{img/supimg_ivgp/2.png}&%
\includegraphics[width=\mywidth,height=\mywidth]{img/supimg_ivgp/woobj/gap/2.png}&\hspace{-4mm}%
\includegraphics[width=\mywidth,height=\mywidth]{img/supimg_ivgp/woobj/gmp/2.png}&\hspace{-4mm}%
\includegraphics[width=\mywidth,height=\mywidth]{img/supimg_ivgp/woobj/glenap/2.png}&&\hspace{-4mm}%
\includegraphics[width=\mywidth,height=\mywidth]{img/supimg_ivgp/wobj/gap/2.png}&\hspace{-4mm}%
\includegraphics[width=\mywidth,height=\mywidth]{img/supimg_ivgp/wobj/gmp/2.png}&\hspace{-4mm}%
\includegraphics[width=\mywidth,height=\mywidth]{img/supimg_ivgp/wobj/glenap/2.png}\vspace{-0.5mm}\\
\includegraphics[width=\mywidth,height=\mywidth]{img/supimg_ivgp/3.png}&%
\includegraphics[width=\mywidth,height=\mywidth]{img/supimg_ivgp/woobj/gap/3.png}&\hspace{-4mm}%
\includegraphics[width=\mywidth,height=\mywidth]{img/supimg_ivgp/woobj/gmp/3.png}&\hspace{-4mm}%
\includegraphics[width=\mywidth,height=\mywidth]{img/supimg_ivgp/woobj/glenap/3.png}&&\hspace{-4mm}%
\includegraphics[width=\mywidth,height=\mywidth]{img/supimg_ivgp/wobj/gap/3.png}&\hspace{-4mm}%
\includegraphics[width=\mywidth,height=\mywidth]{img/supimg_ivgp/wobj/gmp/3.png}&\hspace{-4mm}%
\includegraphics[width=\mywidth,height=\mywidth]{img/supimg_ivgp/wobj/glenap/3.png}\vspace{-0.5mm}\\
\end{tabular}
 \caption{Sample viraliency maps for the IVGP dataset.\label{fig:ivgp-1}}
\end{figure*}

\begin{figure*}
\setlength{\mywidth}{2.3cm}
\centering
 \begin{tabular}{cccccccc}
 Original & \multicolumn{3}{c}{No objectness} && \multicolumn{3}{c}{Objectness}\\
 \cmidrule{2-8}
 Image & GAP~\cite{zhou2015learning} & GMP~\cite{oquab2015object} & G\textsc{lena}P && GAP~\cite{zhou2015learning} & GMP~\cite{oquab2015object} & 
G\textsc{lena}P \\
\includegraphics[width=\mywidth,height=\mywidth]{img/supimg_ivgp/25.png}&%
\includegraphics[width=\mywidth,height=\mywidth]{img/supimg_ivgp/woobj/gap/25.png}&\hspace{-4mm}%
\includegraphics[width=\mywidth,height=\mywidth]{img/supimg_ivgp/woobj/gmp/25.png}&\hspace{-4mm}%
\includegraphics[width=\mywidth,height=\mywidth]{img/supimg_ivgp/woobj/glenap/25.png}&&\hspace{-4mm}%
\includegraphics[width=\mywidth,height=\mywidth]{img/supimg_ivgp/wobj/gap/25.png}&\hspace{-4mm}%
\includegraphics[width=\mywidth,height=\mywidth]{img/supimg_ivgp/wobj/gmp/25.png}&\hspace{-4mm}%
\includegraphics[width=\mywidth,height=\mywidth]{img/supimg_ivgp/wobj/glenap/25.png}\vspace{-0.5mm}\\
\includegraphics[width=\mywidth,height=\mywidth]{img/supimg_ivgp/27.png}&%
\includegraphics[width=\mywidth,height=\mywidth]{img/supimg_ivgp/woobj/gap/27.png}&\hspace{-4mm}%
\includegraphics[width=\mywidth,height=\mywidth]{img/supimg_ivgp/woobj/gmp/27.png}&\hspace{-4mm}%
\includegraphics[width=\mywidth,height=\mywidth]{img/supimg_ivgp/woobj/glenap/27.png}&&\hspace{-4mm}%
\includegraphics[width=\mywidth,height=\mywidth]{img/supimg_ivgp/wobj/gap/27.png}&\hspace{-4mm}%
\includegraphics[width=\mywidth,height=\mywidth]{img/supimg_ivgp/wobj/gmp/27.png}&\hspace{-4mm}%
\includegraphics[width=\mywidth,height=\mywidth]{img/supimg_ivgp/wobj/glenap/27.png}\vspace{-0.5mm}\\
\end{tabular}
 \caption{Sample viraliency maps for the IVGP dataset.\label{fig:ivgp-2}}
\end{figure*}

\begin{figure*}
\setlength{\mywidth}{2.3cm}
\centering
 \begin{tabular}{cccccccc}
 Original & \multicolumn{3}{c}{No objectness} && \multicolumn{3}{c}{Objectness}\\
 \cmidrule{2-8}
 Image & GAP~\cite{zhou2015learning} & GMP~\cite{oquab2015object} & G\textsc{lena}P && GAP~\cite{zhou2015learning} & GMP~\cite{oquab2015object} & 
G\textsc{lena}P \\
\includegraphics[width=\mywidth,height=\mywidth]{img/supimg_ivgp/46.png}&%
\includegraphics[width=\mywidth,height=\mywidth]{img/supimg_ivgp/woobj/gap/46.png}&\hspace{-4mm}%
\includegraphics[width=\mywidth,height=\mywidth]{img/supimg_ivgp/woobj/gmp/46.png}&\hspace{-4mm}%
\includegraphics[width=\mywidth,height=\mywidth]{img/supimg_ivgp/woobj/glenap/46.png}&&\hspace{-4mm}%
\includegraphics[width=\mywidth,height=\mywidth]{img/supimg_ivgp/wobj/gap/46.png}&\hspace{-4mm}%
\includegraphics[width=\mywidth,height=\mywidth]{img/supimg_ivgp/wobj/gmp/46.png}&\hspace{-4mm}%
\includegraphics[width=\mywidth,height=\mywidth]{img/supimg_ivgp/wobj/glenap/46.png}\vspace{-0.5mm}\\
\includegraphics[width=\mywidth,height=\mywidth]{img/supimg_ivgp/52.png}&%
\includegraphics[width=\mywidth,height=\mywidth]{img/supimg_ivgp/woobj/gap/52.png}&\hspace{-4mm}%
\includegraphics[width=\mywidth,height=\mywidth]{img/supimg_ivgp/woobj/gmp/52.png}&\hspace{-4mm}%
\includegraphics[width=\mywidth,height=\mywidth]{img/supimg_ivgp/woobj/glenap/52.png}&&\hspace{-4mm}%
\includegraphics[width=\mywidth,height=\mywidth]{img/supimg_ivgp/wobj/gap/52.png}&\hspace{-4mm}%
\includegraphics[width=\mywidth,height=\mywidth]{img/supimg_ivgp/wobj/gmp/52.png}&\hspace{-4mm}%
\includegraphics[width=\mywidth,height=\mywidth]{img/supimg_ivgp/wobj/glenap/52.png}\vspace{-0.5mm}\\
\end{tabular}
 \caption{Sample viraliency maps for the IVGP dataset.\label{fig:ivgp-3}}
\end{figure*}

\begin{figure*}
\setlength{\mywidth}{2.3cm}
\centering
 \begin{tabular}{cccccccc}
 Original & \multicolumn{3}{c}{No objectness} && \multicolumn{3}{c}{Objectness}\\
 \cmidrule{2-8}
 Image & GAP~\cite{zhou2015learning} & GMP~\cite{oquab2015object} & G\textsc{lena}P && GAP~\cite{zhou2015learning} & GMP~\cite{oquab2015object} & 
G\textsc{lena}P \\
\includegraphics[width=\mywidth,height=\mywidth]{img/supimg_ivgp/72.png}&%
\includegraphics[width=\mywidth,height=\mywidth]{img/supimg_ivgp/woobj/gap/72.png}&\hspace{-4mm}%
\includegraphics[width=\mywidth,height=\mywidth]{img/supimg_ivgp/woobj/gmp/72.png}&\hspace{-4mm}%
\includegraphics[width=\mywidth,height=\mywidth]{img/supimg_ivgp/woobj/glenap/72.png}&&\hspace{-4mm}%
\includegraphics[width=\mywidth,height=\mywidth]{img/supimg_ivgp/wobj/gap/72.png}&\hspace{-4mm}%
\includegraphics[width=\mywidth,height=\mywidth]{img/supimg_ivgp/wobj/gmp/72.png}&\hspace{-4mm}%
\includegraphics[width=\mywidth,height=\mywidth]{img/supimg_ivgp/wobj/glenap/72.png}\vspace{-0.5mm}\\
\includegraphics[width=\mywidth,height=\mywidth]{img/supimg_ivgp/75.png}&%
\includegraphics[width=\mywidth,height=\mywidth]{img/supimg_ivgp/woobj/gap/75.png}&\hspace{-4mm}%
\includegraphics[width=\mywidth,height=\mywidth]{img/supimg_ivgp/woobj/gmp/75.png}&\hspace{-4mm}%
\includegraphics[width=\mywidth,height=\mywidth]{img/supimg_ivgp/woobj/glenap/75.png}&&\hspace{-4mm}%
\includegraphics[width=\mywidth,height=\mywidth]{img/supimg_ivgp/wobj/gap/75.png}&\hspace{-4mm}%
\includegraphics[width=\mywidth,height=\mywidth]{img/supimg_ivgp/wobj/gmp/75.png}&\hspace{-4mm}%
\includegraphics[width=\mywidth,height=\mywidth]{img/supimg_ivgp/wobj/glenap/75.png}\vspace{-0.5mm}\\
\end{tabular}
 \caption{Sample viraliency maps for the IVGP dataset.\label{fig:ivgp-4}}
\end{figure*}

\begin{figure*}
\setlength{\mywidth}{2.3cm}
\centering
 \begin{tabular}{cccccccc}
 Original & \multicolumn{3}{c}{No objectness} && \multicolumn{3}{c}{Objectness}\\
 \cmidrule{2-8}
 Image & GAP~\cite{zhou2015learning} & GMP~\cite{oquab2015object} & G\textsc{lena}P && GAP~\cite{zhou2015learning} & GMP~\cite{oquab2015object} & 
G\textsc{lena}P \\
\includegraphics[width=\mywidth,height=\mywidth]{img/supimg_ivgp/112.png}&%
\includegraphics[width=\mywidth,height=\mywidth]{img/supimg_ivgp/woobj/gap/112.png}&\hspace{-4mm}%
\includegraphics[width=\mywidth,height=\mywidth]{img/supimg_ivgp/woobj/gmp/112.png}&\hspace{-4mm}%
\includegraphics[width=\mywidth,height=\mywidth]{img/supimg_ivgp/woobj/glenap/112.png}&&\hspace{-4mm}%
\includegraphics[width=\mywidth,height=\mywidth]{img/supimg_ivgp/wobj/gap/112.png}&\hspace{-4mm}%
\includegraphics[width=\mywidth,height=\mywidth]{img/supimg_ivgp/wobj/gmp/112.png}&\hspace{-4mm}%
\includegraphics[width=\mywidth,height=\mywidth]{img/supimg_ivgp/wobj/glenap/112.png}\vspace{-0.5mm}\\
\includegraphics[width=\mywidth,height=\mywidth]{img/supimg_ivgp/119.png}&%
\includegraphics[width=\mywidth,height=\mywidth]{img/supimg_ivgp/woobj/gap/119.png}&\hspace{-4mm}%
\includegraphics[width=\mywidth,height=\mywidth]{img/supimg_ivgp/woobj/gmp/119.png}&\hspace{-4mm}%
\includegraphics[width=\mywidth,height=\mywidth]{img/supimg_ivgp/woobj/glenap/119.png}&&\hspace{-4mm}%
\includegraphics[width=\mywidth,height=\mywidth]{img/supimg_ivgp/wobj/gap/119.png}&\hspace{-4mm}%
\includegraphics[width=\mywidth,height=\mywidth]{img/supimg_ivgp/wobj/gmp/119.png}&\hspace{-4mm}%
\includegraphics[width=\mywidth,height=\mywidth]{img/supimg_ivgp/wobj/glenap/119.png}\vspace{-0.5mm}\\
\end{tabular}
 \caption{Sample viraliency maps for the IVGP dataset.\label{fig:ivgp-5}}
\end{figure*}

\begin{figure*}
\setlength{\mywidth}{2.3cm}
\centering
 \begin{tabular}{cccccccc}
 Original & \multicolumn{3}{c}{No objectness} && \multicolumn{3}{c}{Objectness}\\
 \cmidrule{2-8}
 Image & GAP~\cite{zhou2015learning} & GMP~\cite{oquab2015object} & G\textsc{lena}P && GAP~\cite{zhou2015learning} & GMP~\cite{oquab2015object} & 
G\textsc{lena}P \\
\includegraphics[width=\mywidth,height=\mywidth]{img/supimg_uiv/6.png}&%
\includegraphics[width=\mywidth,height=\mywidth]{img/supimg_uiv/woobj/gap/6.png}&\hspace{-4mm}%
\includegraphics[width=\mywidth,height=\mywidth]{img/supimg_uiv/woobj/gmp/6.png}&\hspace{-4mm}%
\includegraphics[width=\mywidth,height=\mywidth]{img/supimg_uiv/woobj/glenap/6.png}&&\hspace{-4mm}%
\includegraphics[width=\mywidth,height=\mywidth]{img/supimg_uiv/wobj/gap/6.png}&\hspace{-4mm}%
\includegraphics[width=\mywidth,height=\mywidth]{img/supimg_uiv/wobj/gmp/6.png}&\hspace{-4mm}%
\includegraphics[width=\mywidth,height=\mywidth]{img/supimg_uiv/wobj/glenap/6.png}\vspace{-0.5mm}\\
\includegraphics[width=\mywidth,height=\mywidth]{img/supimg_uiv/9.png}&%
\includegraphics[width=\mywidth,height=\mywidth]{img/supimg_uiv/woobj/gap/9.png}&\hspace{-4mm}%
\includegraphics[width=\mywidth,height=\mywidth]{img/supimg_uiv/woobj/gmp/9.png}&\hspace{-4mm}%
\includegraphics[width=\mywidth,height=\mywidth]{img/supimg_uiv/woobj/glenap/9.png}&&\hspace{-4mm}%
\includegraphics[width=\mywidth,height=\mywidth]{img/supimg_uiv/wobj/gap/9.png}&\hspace{-4mm}%
\includegraphics[width=\mywidth,height=\mywidth]{img/supimg_uiv/wobj/gmp/9.png}&\hspace{-4mm}%
\includegraphics[width=\mywidth,height=\mywidth]{img/supimg_uiv/wobj/glenap/9.png}\vspace{-0.5mm}\\
\end{tabular}
 \caption{Sample viraliency maps for the UIV dataset.\label{fig:uiv-1}}
\end{figure*}

\begin{figure*}
\setlength{\mywidth}{2.3cm}
\centering
 \begin{tabular}{cccccccc}
 Original & \multicolumn{3}{c}{No objectness} && \multicolumn{3}{c}{Objectness}\\
 \cmidrule{2-8}
 Image & GAP~\cite{zhou2015learning} & GMP~\cite{oquab2015object} & G\textsc{lena}P && GAP~\cite{zhou2015learning} & GMP~\cite{oquab2015object} & 
G\textsc{lena}P \\
\includegraphics[width=\mywidth,height=\mywidth]{img/supimg_uiv/20.png}&%
\includegraphics[width=\mywidth,height=\mywidth]{img/supimg_uiv/woobj/gap/20.png}&\hspace{-4mm}%
\includegraphics[width=\mywidth,height=\mywidth]{img/supimg_uiv/woobj/gmp/20.png}&\hspace{-4mm}%
\includegraphics[width=\mywidth,height=\mywidth]{img/supimg_uiv/woobj/glenap/20.png}&&\hspace{-4mm}%
\includegraphics[width=\mywidth,height=\mywidth]{img/supimg_uiv/wobj/gap/20.png}&\hspace{-4mm}%
\includegraphics[width=\mywidth,height=\mywidth]{img/supimg_uiv/wobj/gmp/20.png}&\hspace{-4mm}%
\includegraphics[width=\mywidth,height=\mywidth]{img/supimg_uiv/wobj/glenap/20.png}\vspace{-0.5mm}\\
\includegraphics[width=\mywidth,height=\mywidth]{img/supimg_uiv/21.png}&%
\includegraphics[width=\mywidth,height=\mywidth]{img/supimg_uiv/woobj/gap/21.png}&\hspace{-4mm}%
\includegraphics[width=\mywidth,height=\mywidth]{img/supimg_uiv/woobj/gmp/21.png}&\hspace{-4mm}%
\includegraphics[width=\mywidth,height=\mywidth]{img/supimg_uiv/woobj/glenap/21.png}&&\hspace{-4mm}%
\includegraphics[width=\mywidth,height=\mywidth]{img/supimg_uiv/wobj/gap/21.png}&\hspace{-4mm}%
\includegraphics[width=\mywidth,height=\mywidth]{img/supimg_uiv/wobj/gmp/21.png}&\hspace{-4mm}%
\includegraphics[width=\mywidth,height=\mywidth]{img/supimg_uiv/wobj/glenap/21.png}\vspace{-0.5mm}\\
\end{tabular}
 \caption{Sample viraliency maps for the UIV dataset.\label{fig:uiv-2}}
\end{figure*}

\begin{figure*}
\setlength{\mywidth}{2.3cm}
\centering
 \begin{tabular}{cccccccc}
 Original & \multicolumn{3}{c}{No objectness} && \multicolumn{3}{c}{Objectness}\\
 \cmidrule{2-8}
 Image & GAP~\cite{zhou2015learning} & GMP~\cite{oquab2015object} & G\textsc{lena}P && GAP~\cite{zhou2015learning} & GMP~\cite{oquab2015object} & 
G\textsc{lena}P \\
\includegraphics[width=\mywidth,height=\mywidth]{img/supimg_uiv/41.png}&%
\includegraphics[width=\mywidth,height=\mywidth]{img/supimg_uiv/woobj/gap/41.png}&\hspace{-4mm}%
\includegraphics[width=\mywidth,height=\mywidth]{img/supimg_uiv/woobj/gmp/41.png}&\hspace{-4mm}%
\includegraphics[width=\mywidth,height=\mywidth]{img/supimg_uiv/woobj/glenap/41.png}&&\hspace{-4mm}%
\includegraphics[width=\mywidth,height=\mywidth]{img/supimg_uiv/wobj/gap/41.png}&\hspace{-4mm}%
\includegraphics[width=\mywidth,height=\mywidth]{img/supimg_uiv/wobj/gmp/41.png}&\hspace{-4mm}%
\includegraphics[width=\mywidth,height=\mywidth]{img/supimg_uiv/wobj/glenap/41.png}\vspace{-0.5mm}\\
\includegraphics[width=\mywidth,height=\mywidth]{img/supimg_uiv/49.png}&%
\includegraphics[width=\mywidth,height=\mywidth]{img/supimg_uiv/woobj/gap/49.png}&\hspace{-4mm}%
\includegraphics[width=\mywidth,height=\mywidth]{img/supimg_uiv/woobj/gmp/49.png}&\hspace{-4mm}%
\includegraphics[width=\mywidth,height=\mywidth]{img/supimg_uiv/woobj/glenap/49.png}&&\hspace{-4mm}%
\includegraphics[width=\mywidth,height=\mywidth]{img/supimg_uiv/wobj/gap/49.png}&\hspace{-4mm}%
\includegraphics[width=\mywidth,height=\mywidth]{img/supimg_uiv/wobj/gmp/49.png}&\hspace{-4mm}%
\includegraphics[width=\mywidth,height=\mywidth]{img/supimg_uiv/wobj/glenap/49.png}\vspace{-0.5mm}\\
\end{tabular}
 \caption{Sample viraliency maps for the UIV dataset.\label{fig:uiv-3}}
\end{figure*}

\begin{figure*}
\setlength{\mywidth}{2.3cm}
\centering
 \begin{tabular}{cccccccc}
 Original & \multicolumn{3}{c}{No objectness} && \multicolumn{3}{c}{Objectness}\\
 \cmidrule{2-8}
 Image & GAP~\cite{zhou2015learning} & GMP~\cite{oquab2015object} & G\textsc{lena}P && GAP~\cite{zhou2015learning} & GMP~\cite{oquab2015object} & 
G\textsc{lena}P \\
\includegraphics[width=\mywidth,height=\mywidth]{img/supimg_uiv/67.png}&%
\includegraphics[width=\mywidth,height=\mywidth]{img/supimg_uiv/woobj/gap/67.png}&\hspace{-4mm}%
\includegraphics[width=\mywidth,height=\mywidth]{img/supimg_uiv/woobj/gmp/67.png}&\hspace{-4mm}%
\includegraphics[width=\mywidth,height=\mywidth]{img/supimg_uiv/woobj/glenap/67.png}&&\hspace{-4mm}%
\includegraphics[width=\mywidth,height=\mywidth]{img/supimg_uiv/wobj/gap/67.png}&\hspace{-4mm}%
\includegraphics[width=\mywidth,height=\mywidth]{img/supimg_uiv/wobj/gmp/67.png}&\hspace{-4mm}%
\includegraphics[width=\mywidth,height=\mywidth]{img/supimg_uiv/wobj/glenap/67.png}\vspace{-0.5mm}\\
\includegraphics[width=\mywidth,height=\mywidth]{img/supimg_uiv/72.png}&%
\includegraphics[width=\mywidth,height=\mywidth]{img/supimg_uiv/woobj/gap/72.png}&\hspace{-4mm}%
\includegraphics[width=\mywidth,height=\mywidth]{img/supimg_uiv/woobj/gmp/72.png}&\hspace{-4mm}%
\includegraphics[width=\mywidth,height=\mywidth]{img/supimg_uiv/woobj/glenap/72.png}&&\hspace{-4mm}%
\includegraphics[width=\mywidth,height=\mywidth]{img/supimg_uiv/wobj/gap/72.png}&\hspace{-4mm}%
\includegraphics[width=\mywidth,height=\mywidth]{img/supimg_uiv/wobj/gmp/72.png}&\hspace{-4mm}%
\includegraphics[width=\mywidth,height=\mywidth]{img/supimg_uiv/wobj/glenap/72.png}\vspace{-0.5mm}\\
\end{tabular}
 \caption{Sample viraliency maps for the UIV dataset.\label{fig:uiv-4}}
\end{figure*}

{\small
\bibliographystyle{ieee}
\bibliography{viraliency.bib}
}
